%% file: main.tex
\documentclass[11pt,a4paper]{article}
\usepackage{authblk}
\usepackage[hyperref]{naaclhlt2019}
\usepackage{times}
\usepackage{latexsym}
\usepackage{graphicx}
\usepackage{amsmath}
\usepackage{afterpage}
\usepackage{url}
\usepackage{enumitem}
\usepackage{amssymb}
\usepackage{algorithm}
\usepackage{breqn}
\usepackage{algpseudocode}
\usepackage{float}
\usepackage{mathtools}
\newlength\myindent
\newcommand{\data}{{MathQA}}
\newcommand{\aqua}{{AQuA}}


\aclfinalcopy 
\def\aclpaperid{2084} 

\title{MathQA: Towards Interpretable Math Word Problem Solving \\ with Operation-Based Formalisms}

\author[1]{Aida Amini}
\author[1]{Saadia Gabriel}
\author[1]{Shanchuan Lin}
\author[1]{Rik Koncel-Kedziorski}  
\author[1,2]{\\Yejin Choi}
\author[1,2]{Hannaneh Hajishirzi}
\affil[1]{University of Washington}
\affil[2]{Allen Institute for AI}
\affil[ ]{\textit {\{amini91, skgabrie, linsh, kedzior, yejin, hannaneh\}@cs.washington.edu}}

\date{}

\begin{document}
\maketitle

\begin{abstract}
We introduce a large-scale dataset of math word problems and an interpretable neural math problem solver that learns to map problems to operation programs.  
Due to annotation challenges, current datasets in this domain have been either relatively small in scale or did not offer precise operational annotations over diverse problem types. We introduce a new representation language to model precise operation programs corresponding to each math problem that aim to improve both the performance and the interpretability of the learned models. Using this representation language, our new dataset, MathQA, significantly enhances the \aqua\ dataset with fully-specified operational programs. We additionally introduce a neural sequence-to-program model enhanced with automatic problem categorization. Our experiments show improvements over competitive baselines in our MathQA as well as the \aqua\ datasets. The results are still significantly lower than human performance indicating that the dataset poses new challenges for future research. Our dataset is available at: \url{https://math-qa.github.io/math-QA/}.
\end{abstract}

\input{intro}

\input{related}
\input{language}
\input{data}

\input{models}
\input{experiments}

\input{conclusion}
\bibliography{naaclhlt2019}
\bibliographystyle{acl_natbib}

\end{document}

%% file: intro.tex
\section{Introduction} 
\begin{figure}[t]
\centering
  \includegraphics[width=.50\textwidth]{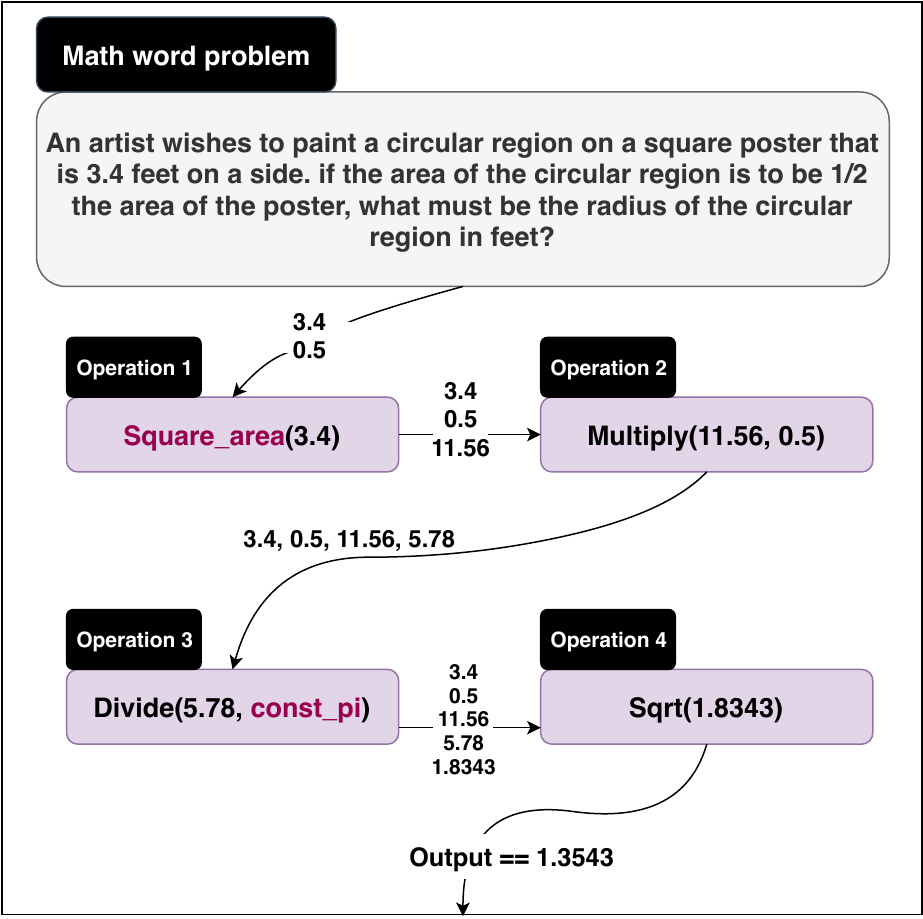}
 \setlength{\belowcaptionskip}{-20pt} 
 \caption{Example of a  math word problem aligned with representation language by crowd-sourced annotation }
    \label{fig:rep}
\end{figure}
 Answering math word problems poses unique challenges for logical reasoning over implicit or explicit quantities expressed in text. Math word-problem solving requires extraction of salient information from 
 natural language narratives. 
 Automatic solvers must transform the textual narratives into executable meaning representations, a process that requires both high precision and, in the case of story problems, significant world knowledge. 
 
 As shown by the geometry question in Figure \ref{fig:rep}, math word problems are generally narratives describing the progress of actions and relations over some entities and quantities. The operation program underlying the problem in Figure \ref{fig:rep} highlights the complexity of the problem-solving task.
 \space{}Here, we need the ability to deduce implied constants (pi) and knowledge of domain-specific formulas (area of the square).

In this paper, we introduce a new operation-based representation language for solving math word problems. 
We use this representation language to construct \data \footnote{The dataset is available at: \url{https://math-qa.github.io/math-QA/}}, a new large-scale, diverse dataset of 37k English multiple-choice math word problems covering multiple math domain categories by modeling operation programs corresponding to word problems in the \aqua{ }dataset \cite{aqua}. We introduce a neural model for mapping problems to operation programs with domain categorization. 

Most current datasets in this domain are small in scale~\cite{alg514} or do not offer precise operational annotations over diverse problem types~\cite{aqua}. This is mainly due to the fact that annotating  math word problems precisely across diverse problem categories is  challenging even for humans, requiring background math knowledge for annotators. Our representation language facilitates the annotation task for crowd-sourcing and increases the interpretability of the proposed model. 

 Our sequence-to-program model with categorization trained on our \data\ dataset outperforms previous state-of-the-art on the \aqua\ test set in spite of the smaller training size. These results indicate the superiority of our representation language and the quality of the formal annotations in our dataset. Our model achieves competitive results on \data, but is still lower than  human  performance  indicating  that  the dataset  poses  new  challenges  for  future  research. 
Our contributions are as follows:
\begin{itemize}[topsep=2pt, leftmargin=4mm, noitemsep]
\item We introduce a large-scale dataset of math word problems that are densely annotated with operation programs
\item We introduce a new representation language to model operation programs corresponding to each math problem that aim to improve both the performance and the interpretability of the learned models. 
\item We introduce a neural architecture leveraging a sequence-to-program model with automatic problem categorization, achieving competitive results on our dataset as well as the \aqua\ dataset
\end{itemize}

%% file: related.tex
\section{Background and Related Work} 
\noindent \textbf{Large-Scale Datasets}  
Several large-scale math word problem datasets have been released in recent years. These include Dolphin18K \cite{dolphin18k}, Math23K \cite{math23k} and \aqua.
We choose the 2017 AQUA-RAT dataset to demonstrate use of our representation language on an existing large-scale math word problem solving dataset. The \aqua\ provides over 100K GRE- and GMAT-level math word problems. The problems are multiple choice and come from a wide range of domains. 

The scale and diversity of this dataset makes it particularly suited for use in training deep-learning models to solve word problems. However there is a significant amount of unwanted noise in the dataset, including problems with incorrect solutions, problems that are unsolvable without brute-force enumeration of solutions, and rationales that contain few or none of the steps required to solve the corresponding problem. The motivation for our dataset comes from the fact we want to maintain the challenging nature of the problems included in the \aqua\ dataset, while removing noise that hinders the ability of neuralized models to learn the types of signal neccessary for problem-solving by logical reasoning.

\noindent \textbf{Additional Datasets}  
Several smaller datasets have been compiled in recent years. Most of these works have focused on algebra word problems, including MaWPS \cite{mawps}, Alg514 \cite{alg514}, and DRAW-1K \cite{draw1k}. Many of these datasets have sought to align underlying equations or systems of equations with word problem text. While recent works like \cite{statsolver, regexp} have explored representing math word problems with logical formalisms and regular expressions, our work is the first to provide well-defined formalisms for representing intermediate problem-solving steps that are shown to be generalizable beyond algebra problems.
 
\vspace{.2cm}
\noindent \textbf{Solving with Handcrafted Features}  Due to sparsity of suitable data, early work on math word problem solving used pattern-matching to map word problems to mathematical expressions \cite{early1,early2,early3}, as well as non-neural statistical modeling and semantic parsing approaches \cite{early4}. 

Some effort has been made on parsing the problems to extract salient entities \cite{verbcat}. This approach views entities as containers, which can be composed into an equation tree representation. The equation tree representation is changed over time by operations implied by the problem text. 

Many early works focused on solving addition and subtraction problems \cite{add1,add2,add3}. As word problems become more diverse and complex, we require models capable of solving simultaneous equation systems. This has led to an increasing focus on finding semantic alignment of math word problems and mentions of numbers \cite{map2declare}. The main idea behind those work is to find all possible patterns of equations and rank them based on the problem. 

\vspace{.2cm}
\noindent \textbf{Neural Word Problem Solvers}  Following the increasing availability of large-scale datasets like \aqua, several recent works have explored deep neural approaches to math word problem solving \cite{math23k}. Our representation language is motivated by exploration of using intermediate formalisms in the training of deep neural problem-solving networks, as is done in the work of \cite{interrep} to solve problems with sequence to sequence models. While this work focused on single-variable arithmetic problems, our work introduces a formal language of operations for covering more complex multivariate problems and systems of equations. 

\vspace{.2cm}
\noindent \textbf{Interpretability of Solvers}  While the statistical models with handcrafted features introduced by prior work are arguably ``interpretable" due to the relative sparsity of features as well as the clear alignments between inputs and outputs, new neuralized approaches present new challenges to model interpretability of math word problem solvers \cite{reinforcement}. While this area is relatively unexplored, a prior approach to increasing robustness and interpretability of math word problem-solving models looks at using an adversarial dataset to determine if models are learning logical reasoning or exploiting dataset biases through pattern-matching \cite{statsolver}. 

%% file: language.tex
\section{Representing Math Word Problems} \label{sec:lang}

\begin{figure}
    \centering
     \setlength{\belowcaptionskip}{-10pt} 
    \includegraphics[width=.4\textwidth]{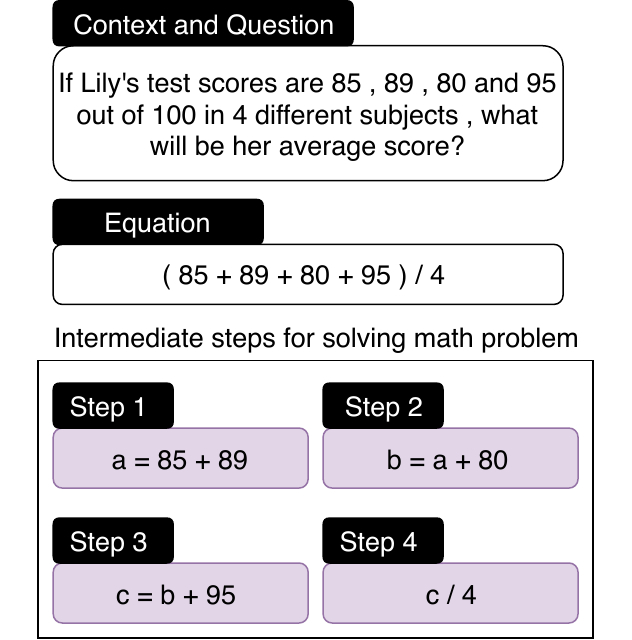}
    \caption{Example of a math word problem with its underlying equation and intermediate steps for problem-solving}
    \label{fig:running_example}
\end{figure}

A math word problem consists of a narrative that grounds mathematical formalisms in real-world concepts. Solving these problems is a challenge for both humans and automatic methods like neural network-based solvers, since it requires logical reasoning about implied actions and relations between entities. For example, in Figure \ref{fig:running_example}, operations like addition and division are not explicitly mentioned in the word problem text, but they are implied by the question. 
As we examine the context of a math word problem, we have to select arguments for operations based on which values are unimportant for solving the problem and which are salient. In Figure \ref{fig:running_example}, the numeric value ``100" appears in the context but does not appear in the underlying equation. 

By selecting implied operations and arguments, we can generate a program of intermediate steps for solving a math word problem. Each step involves a mathematical operation and its related arguments. In Figure 2, there are three addition operations and one division. As illustrated in the figure, operations can be dependant to the previous ones by the values they use as arguments. Every math word problem can be solved by sequentially executing these programs of dependent operations and arguments.  
 
 We define formalisms for expressing these sequential operation programs with a domain-aware representation language. An operation program in our representation language is a sequence with \(n\) operations. The general form is shown below. Each operation \(o_i\) takes in a list of arguments \textbf{a} of length \(i\):
\begin{equation}
o_1(\textbf{a}_1) o_2(\textbf{a}_2)  ... o_n(\textbf{a}_n) 
\end{equation}
Given this general definition, the problem in Figure \ref{fig:running_example} has the following representation\footnote{Here the arguments \(174\), \(254\) and \(349\) are the outputs of operations 1, 2 and 3 respectively.}:
\begin{equation}
\begin{split}
\text{add}_1( 85 , 89 ) \text{add}_2( 174 , 80 ) \\
\text{add}_3( 254 , 95 ) \text{divide}_4( 349 , 4 ) 
\end{split}
\end{equation}

Our representation language consists of 58 operations and is designed considering the following objectives. 
\begin{itemize}
    \item Correctness \(\rightarrow\) Operation programs should result in the correct solution when all operations are executed.
    \item Domain-awareness \(\rightarrow\) Operation problems should make use of both math knowledge and domain knowledge associated with subfields like  geometry and probability to determine which operations and arguments to use.
    \item Human interpretability \(\rightarrow\) Each operation and argument used to obtain the correct solution should relate to part of the input word problem context or a previous step in the operation program.
\end{itemize}
Learning logical forms has led to success in other areas of semantic parsing \cite{semantic1,semantic2,semantic3,semantic4} and is a natural representation for math word problem-solving steps. By augmenting our dataset with these formalisms, we are able to cover most types of math word problems\footnote{We omit high-order polynomials and problems where the solutions are entirely nonnumeric.}. In contrast to other representations like simultaneous equations, our formalisms ensure that every problem-solving step is aligned to a previous one. There are three advantages to this approach. First, we use this representation language to provide human annotators with clear steps for how a particular problem should be solved with math and domain knowledge. Second, our formalisms provide neural models with a continuous path to execute operations for problems with systems of equations, instead of forcing models to align equations before problem solving. This reduces the possibility of intermediate errors being propagated and leading to a incorrect solution. 
Finally, by having neural models generate a solution path in our representation language before computing the final solution, we are able to reconstruct the logical hops inferred by the model output, increasing model interpretability. 


%% file: data.tex
\section{Dataset}\label{sec:data}
\begin{figure*}
\centering
\setlength{\belowcaptionskip}{-12pt}
  \includegraphics[width=1\textwidth]{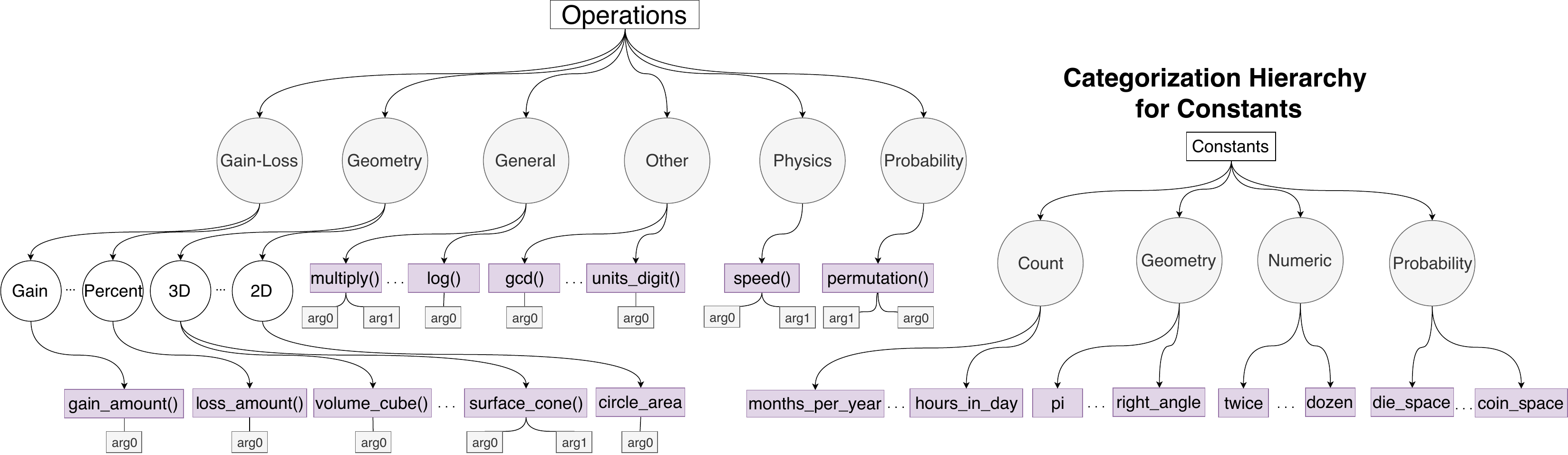}
 \caption{Category-based Hierarchies for Operation Formalisms }
    \label{fig:categorization}
\end{figure*}
Our dataset (called~\data) consists of 37,200 math word problems, corresponding lists of multiple-choice options and aligned operation programs. 
We use problems in the \aqua\ dataset and carefully annotate those problems with formal operation programs. 


Math problems are first categorized into math domains using term frequencies (more details in Section~\ref{sec:cat_setup}). These domains are used to prune the search space of possible operations to align with the word problem text.
Figure \ref{fig:categorization} shows the category-based hierarchies for operation formalisms. 

We use crowdsourcing to carefully align problems with operation programs (Section~\ref{sec:crowd}). Table~\ref{table:datastats} shows overall statistics of the dataset.\footnote{We also experimented with an automatic dynamic programming approach to annotation that generates operation programs for problems using numbers in the \aqua\ rationales. Due to the noise in the rationales, only $61\%$ of those problems pass our human validation. 
This is mainly due to the fact that the rationales are not complete programs and fail to explicitly describe all important numbers and operations required to solve the problem. To maintain interpretability of operation paths, we did not include automatic annotations from our dataset and focus on operation programs derived by crowdsourcing. }

\begin{table}[t]
\resizebox{\columnwidth}{!}{%
\begin{tabular}{  l llll } 
Category  & \#Prob. &  Avg \#words & \#Vocab  & Avg \#ops  \\ \hline
Geometry  & 3,316 & 34.3 & 1,839 &  4.8  \\ 
Physics & 9,830 & 37.3 & 3,340 & 5.0 \\
Probability  & 663  & 38.9 & 937 &  5.0  \\
Gain-Loss  &  4,377 & 34.3 & 1,533 &  5.7 \\ 
General  & 17,796 & 38.6 & 6,912 & 5.1 \\ 
Other  &  1,277 & 31.3 &  1,425 & 4.7 \\ \hline
All & 37,259 & 37.9 & 6,664 & 5.3 \\
\end{tabular}}
\caption{Statistics for our dataset; the total number of operations in the dataset is 58.}
\label{table:datastats}
\end{table}


\subsection{Annotation using Crowd Workers}\label{sec:crowd}
 Annotating GRE level math problems can be a challenging and time consuming task for humans. We design a dynamic annotation platform to annotate math word problems with formal operation programs.   Our annotation platform has the following properties: (a) it provides basic math knowledge to annotators, (b) it is dynamic by iteratively calculating intermediate results after an operation submission, and (c) it employs quality control strategies. 
 
\paragraph{Dynamic Annotation Platform} 
The annotators are provided with a problem description, a list of operations related to the problem category, and a list of valid arguments. They  iteratively select operations and arguments until the problem is solved. 

\begin{itemize}[topsep=2pt, leftmargin=4mm]
\item{\bf  Operation Selection} The annotators are instructed to sequentially select an operation from the list of operations in the problem category. Annotators are provided with math  knowledge  by hovering over every operation and getting the related hint that consists of arguments, formula and a short explanation of the operation. 
 \item{\bf  Argument Selection}
 After selecting the operation the list of valid arguments are presented to the annotators to choose from. Valid arguments consist of numbers in the problem, constants in the problem category, and the previous calculations.  The annotators are restricted to select only from these valid arguments to prevent having noisy and dangling numbers.
 After submission of an operation and the corresponding arguments, the result of the operation is automatically calculated and will be added as a new valid argument to the argument list.    
 \item{\bf Program Submission}
To prevent annotators from submitting arbitrary programs,  we enforce restrictions to the final submission. Our platform only accepts programs which include some numbers from the problem, and whose final calculation is very close to the correct numeric solution. 
\end{itemize}
\paragraph{High Quality Crowd Workers} We dynamically evaluate and employ high-quality annotators through a collection of quality-control questions. We take advantage of the annotation platform in \textit{Figure Eight}.\footnote{https://www.figure-eight.com} The annotators are randomly evaluated through a pre-defined set of test questions, and they have to maintain an accuracy threshold to be able to continue their annotations. If an annotator's accuracy drops below a threshold, their previous annotations are labeled as untrusted and will be added to the pool of annotations again. 

\paragraph{Alignment Validation} To further evaluate the quality of the annotated programs, we leverage a validation strategy to check whether the  problems and annotated programs are aligned or not. According to this strategy, at least 2 out of 3 validators should rank the operation program as valid for it to be selected. The validation accuracy is $94.64\%$ across categories.

%% file: models.tex
\section{Models}
\begin{figure*}
\centering
\setlength{\belowcaptionskip}{-12pt}
  \includegraphics[width=1\textwidth]{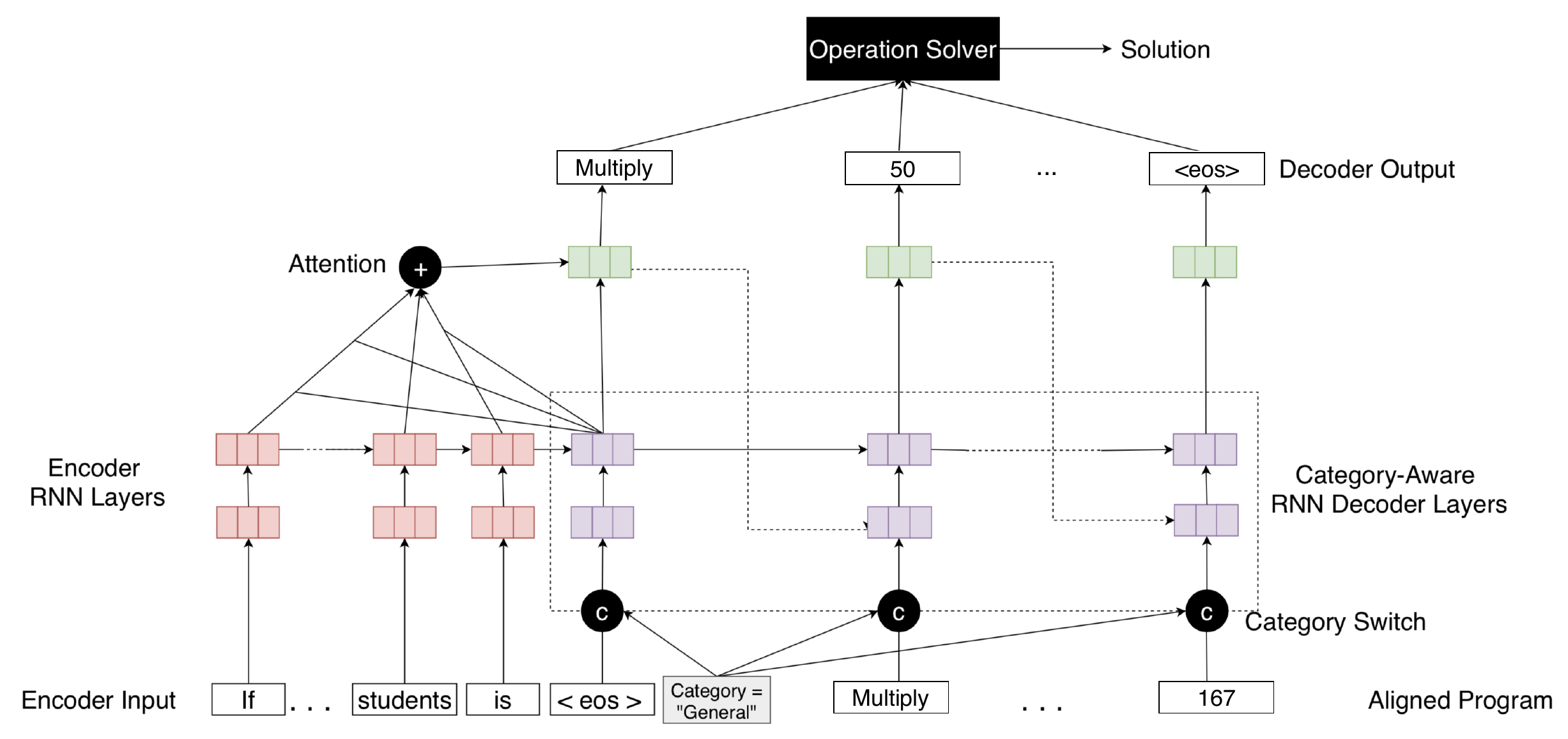}
 \caption{Architecture of sequence-to-program model with categorization. Shown with example problem ``If the average marks of three batches of 62 , 60 and 45 students respectively is 50 , 55 , 60 , then the average marks of all the students is." }
    \label{fig:model}
\end{figure*}
We develop encoder-decoder neural models to map word problems to a set of feasible operation programs.  We match the result of the executed operation program against the list of multiple-choice options given for a particular problem. The matching solution is the final model output. 

We frame the problem of aligning an operation program with a math word problem as a neural machine translation (NMT) task, where the word problem \(\mathbf{x}\) and gold operation program \(\mathbf{y}\) form a parallel text pair. The vocabulary of \(\mathbf{y}\) includes all possible operations and arguments in our representation language. 
\subsection{Sequence-to-Program}
For our initial sequence-to-program model, we follow the attention-based NMT paradigm of \cite{nmt, nmt2}. We encode the source word problem text \(\mathbf{x} = (x_1, x_2, ... , x_M)\) using a bidirectional RNN encoder \(\theta^{enc}\). The decoder \(\theta^{dec}\) predicts a distribution over the vocabulary and input tokens to generate each operation or argument in the target operation program. For our sequence-to-program model vocabulary, we use informed generation, in which the program tokens are generated separately from the vocabulary of operations $O$ or arguments $A$. 

The encoded text is represented by a sequence of $d$-dimensional hidden states \(\mathbf{h}^{enc} = (h^{enc}_1,h^{enc}_2,..,h^{enc}_M)\), where \(M\) is the length of the input text. A context vector $a_i$ is computed by taking the weighted sum of the attention model weights \(\alpha_{t,i}\) for each timestep \(t \in (1,2,..., T)\) and each encoder hidden state \(h^{enc}_i\): \begin{center} $a_i = \sum_{i=1}^{M} \alpha_{t,i}h^{enc}_i$.\end{center}

We compute the $d$-dimensional decoder hidden state \(h^{dec}_i\) using a LSTM recurrent layer:
\begin{equation}\label{eq:2}
h^{dec}_i = LSTM(h^{dec}_{i-1}, y_{i-1},a_i)
\end{equation}
At each timestep, we make a prediction for an operator \(op_i\) or argument \(arg_{ik}\), where \(k\) corresponds to the index of the argument in operator \(i\)'s argument list. This prediction is conditioned on the previous tokens \((y_1,...,y_{i-1})\) and the input \(\mathbf{x}\) to decode an entire operation program \(\mathbf{y}=(y_1,y_2,...,y_N)\) of length $N$: 
\begin{eqnarray}
P(\mathbf{y}| x) = \prod_{i=1}^{N} P(y_i | y_{<i},\mathbf{x}) \label{eq:3} \\
P(y_i | y_{<i}, \mathbf{x}) = g(f(h^{dec}_i, y_i, a_i)) \label{eq:4}
\end{eqnarray}
Here \(f\) is a 1-layer feed-forward neural network and \(g\) is the softmax function. During training time, we minimize the negative log-likelihood (NLL) using the following objective:
\begin{eqnarray}\label{eq:5}
\mathcal{L}(\theta^{enc},\theta^{dec}) = -log P(\mathbf{y}|\mathbf{x};\theta^{enc},\theta^{dec})
\end{eqnarray}
At test time, we only observe the input text when predicting operation programs:
\begin{equation}\label{eq:6}
\hat{\mathbf{y}}=\texttt{argmax}_{\mathbf{y}}P(\mathbf{y}|\mathbf{x})
\end{equation}
\subsection{Categorized Sequence-to-Program Model}
We extend our base sequence-to-program model to integrate knowledge of math word problem domain categories. We modify the RNN decoder layers that compute the decoder hidden state to be category-aware. Here, the category label $c$ is deterministically computed by the category extractor (explained below). It functions as a hard decision switch that determines which set of parameters to use for the hidden state computation: 
\begin{eqnarray}\label{eq:7}
h^{dec}_i = LSTM_{c}(h^{dec}_{i-1}, y_{i-1},a_i)
\end{eqnarray}
The updated objective function from equation (\ref{eq:6}) is shown below:
\begin{equation}\label{eq:8}
\mathcal{L}(\theta^{enc},\theta^{dec}_{c}) = -log P(\mathbf{y}|\mathbf{x};\theta^{enc},\theta^{dec}_{c})
\end{equation}
The full model architecture is shown in Figure \ref{fig:model}. 

\paragraph{Domain-Specific Category Extraction}\label{sec:cat_setup}
We first construct a lexicon of n-grams relating to a specific domain. The lexicon is a list consisting of domain-specific categories and associated n-grams. For each domain category $c$ in the lexicon, we select associated n-grams $\textbf{n}_c$ that occur frequently in word problems belonging to domain category $c$, but rarely appear in other domain categories. We compute n-gram frequency $f_{pc}$ as the number of n-grams associated with a category $c$ appearing in the text of a word problem $p$. We obtain a list of potential categories for $p$ by choosing all categories for which $f_{pc} > 0$, and then assign a category label to $p$ based on which category has the highest n-gram frequency. 
\subsection{Solving Operation Programs}
Once a complete operation program has been decoded, each operator in the program is executed sequentially along with its predicted set of arguments to obtain a possible solution. For each word problem $p$ and options $o$, we generate a beam of the top $n$ decoded operation programs. We execute each decoded program $g$ to find the solution from the list of options $\textbf{o}$ of the problem. We first choose options that are within a threshold of the executed value of $g$. We select $g$ as the predicted solution by checking the number of selected options  and the minimum distance between the executed value of $g$ and a possible option for $p$.
For the problems in  \aqua{}  that do not belong in any category of MathQA, we randomly choose an option.

%% file: experiments.tex
\section{Experimental Setup}\label{sec:setup}
\subsection{Datasets}
Our dataset consists of  $37k$ problems which are randomly split in $(80/12/8)\%$ training/dev/test problems. Our dataset significantly enhances the \aqua\ dataset by  fully annotating a portion of {\it solvable} problems in the \aqua\ dataset into formal operation programs.

We carefully study the \aqua\ dataset. Many of the problems are near-duplicates with slight changes to the math word problem stories or numerical values since they are expanded from a set of 30,000 seed problems through crowdsourcing \cite{aqua}. These changes are not always reflected in the rationales, leading to incorrect solutions. There are also some problems that are not solvable given current math word problem solving frameworks because they require a level of reasoning not yet modeled by neural networks. Sequence problems, for example, require understanding of patterns that are difficult to intuit without domain knowledge like sequence formulas, and can only be solved automatically through brute-force or guessing.  Table \ref{table:solve-table} shows a full breakdown of the \aqua\ dataset by solvability.\footnote{There is overlap between unsolvable subsets. For example, a sequence problem may also be a duplicate of another problem in the \aqua\ dataset.} 
\begin{table}[h]
\resizebox{\columnwidth}{!}{%
\begin{tabular}{ l | l  | l  } 
Subset   & Train & Valid \\ \hline
Unsolvable - No Words &  37  & 0  \\ 
Unsolvable - Sequence & 1,991  & 4   \\ 
Unsolvable - Requires Options &  6,643   & 8 \\ 
Unsolvable - Non-numeric &  10,227   & 14  \\ 
Duplicates  &  17,294  & 0  \\ \hline
Solvable & 65,991 & 229 \\ \hline
Total  &  97,467 & 254 \\
\end{tabular}
}
\caption{Full original AQuA solvability statistics.} 
\label{table:solve-table}
\end{table}

\subsection{Annotation Details}
 We follow the annotation strategy described in Section~\ref{sec:data} to formally annotate problems with operation programs. \footnote{We tried two other strategies of showing extra information (rationales or end solutions) to annotators to facilitate solving problems. However, our manual validation showed that annotators mostly used those extra information to artificially build an operation program without reading the problem.}

\paragraph{Annotator Agreements and Evaluations} Our expert evaluation of the annotation procedure for a collection of 500 problems shows that 92\% of the annotations are valid. Additionally, it has $87\%$ agreement between the expert validation and the crowd sourcing validation task.

\paragraph{Annotation Expansion} The \aqua\ dataset consists of a group of problems which share similar characteristics. These problems can be solved with similar operation programs.  We find closely similar problems, replace numeric values with generic numbers, and expand annotations to cover more problems from the \aqua\ dataset. For similarity, we use Levenshtein distance with a threshold of 4 words in edit distance. 

\begin{figure*}
\setlength{\belowcaptionskip}{-12pt} 
\centering
  \includegraphics[width=.9\textwidth]{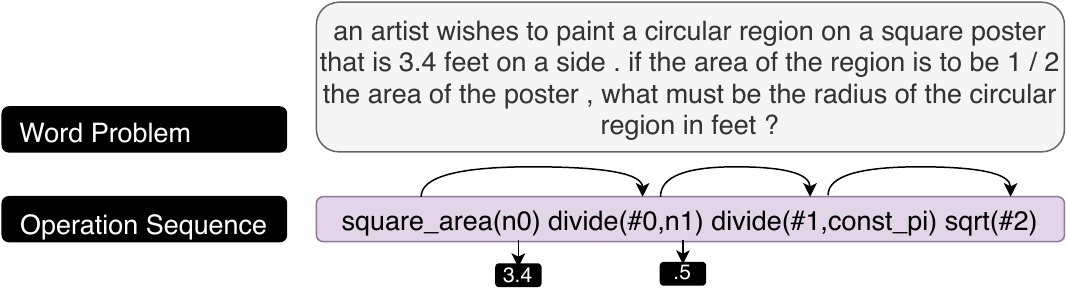}
 \caption{Example of an operation program generated by our Seq2prog model with categorization }
    \label{fig:gen}
\end{figure*}
\subsection{Model and Training Details}
We use the official python implementation of OpenNMT \cite{2017opennmt}. We choose a LSTM-based encoder-decoder architecture. 
We use Adam optimizer~\cite{adam}, and the learning rate for training is $0.001$. The hidden size for the encoder and decoder is set to $d = 100$. Both the encoder and decoder have $2$ layers.  The word embedding vectors are randomly initialized. At inference time, we implemented a beam search with beam size of 200 for \aqua\ and 100 for MathQA. 

 The program vocabulary consists of the operations $O$ in our representation language and valid arguments $A$. For valid arguments, we do not use their actual values since the space is very large. Instead, we keep a list of numbers according to their source. Constants are predefined numbers that are available to all problems. Problem numbers are added to the list according to their order in the problem text. Calculated numbers in the intermediate steps are added to the list according to the operation order. 

\section{Experimental Results}



\subsection{Results} Table \ref{table:results} compares the performance of our sequence-to-program models trained on \data\ with baselines on \data\ and \aqua\ test sets. The base model is referred to as ``Seq2prog," while our model with categorization is ``Seq2prog + cat." For accuracy, the performance was measured in terms of how well the model would perform on an actual math test.

We observe improvement for our ``Seq2prog + cat'' model despite the fact that our training data is proportionally smaller than the \aqua\ dataset,  and our model is much simpler than the state-of-the-art model on this dataset. This indicates the effectiveness of our formal representation language to incorporate domain knowledge as well as the quality of the annotations in our dataset. 


\begin{table}[t]
\centering
\setlength{\belowcaptionskip}{-12pt} 
\begin{tabular}{lll} 
Model  &  \data & \aqua \\ \hline
Random  & 20.0 & 20.0 \\ 
\aqua\ Model & - & 36.4\\
Seq2prog  &  51.9 & 33.0\\
Seq2prog + cat  &  \textbf{54.2} & \textbf{37.9}\\ \hline
\end{tabular}
\caption{Experimental results for accuracy on our \data\ and  \aqua\ test sets}
\label{table:results}
\end{table}

\subsection{Analysis}
\paragraph{Qualitative Analysis }
Table \ref{table:correct_sol} and Figure~\ref{fig:gen} show  some examples of problems solved by
our method. We analyzed 50 problems that are solved wrongly by our system on the MathQA dataset. Table~\ref{table:quantative} summarizes four major categories of errors. 

The most common type of errors are  problems that need complicated or long chain of mathematical reasoning. For example, the first problem in Table~\ref{table:quantative} requires reasoning that goes beyond one sentence. 
Other errors  are due to limitations in our representation language. For example, the second problem in Table~\ref{table:quantative} requires the {\it factorization} operation  which is not defined in our representation language.  Future work can investigate more domains of mathematics such as logic, number factors, etc. Some errors are due to the slightly noisy
nature of our categorization strategy. For example, the third problem in Table~\ref{table:quantative}  is mistakenly categorized as belonging to {\it physics} domain due to the presence of words {\it m, cm, liter} in the problem text, while the correct category for the problem is {\it geometry}. The final category of errors are due to problems that do not have enough textual context or erroneous problems (e.g., fourth problem in Table~\ref{table:quantative}).


\begin{table*}[t]
    \centering
    \resizebox{\textwidth}{!}{%
    \begin{tabular}{|p{5cm}|p{13cm}|}
        \hline
        \textbf{Error type}  & \textbf{Problem}  \\ \hline
        Hard problems ($45\%$)   &  Jane and Ashley take 8 days and 40 days respectively to complete a project when they work on it alone. They thought if they worked on the project together, they would take fewer days to complete it. During the period that they were working together, \underline{Jane took an eight day leave from work}. This led to Jane' s working for four extra days on her own to complete the project. How long did it take to finish the project?            \\ \hline 
        Limitation in representation language  ($25\%$) & How many different positive integers  are \underline{factors} of 25?\\ \hline
        Categorization errors    ($12.5\%$)       & A cistern of capacity 8000 \underline{litres} measures externally 3.3 \underline{m} by 2.6 \underline{m} by 1.3 \underline{m} and its walls are 5 \underline{cm} thick. The thickness of the bottom is:     \\ \hline
        Incorrect or insufficient problem text) ($17.5\%$) & 45 x \underline{?} = 25 $\%$ of 900    \\ \hline
    \end{tabular}
    }
    \caption{Examples of mistakes made by our system. The reason of the errors are underlined. }
    \label{table:quantative}
\end{table*}

\begin{table}[t]
    \centering
    \resizebox{\columnwidth}{!}{%
    \begin{tabular}{|p{8.5cm}|}
        \hline
        \textbf{Problem :} A rectangular field is to be fenced on three sides leaving a side of 20 feet uncovered. if the area of the field is 10 sq. feet, how many feet of fencing will be required? \newline 
        \textbf{Operations :} \texttt{divide(10,20), multiply($\#0$, const_2), add(20, \#1)}\\ \hline
        \textbf{Problem :} How long does a train 110m long running at the speed of 72 km/hr takes to cross a bridge 132m length?  \newline
        \textbf{Operations :} \texttt{add(110, 132), multiply(72, const_0.2778), divide($\#0$, $\#1$), floor($\#2$)}  \\ \hline
    
    \end{tabular}
    }
    \caption{Problems solved correctly by Seq2prog+cat model.}
    \label{table:correct_sol}
\end{table}

\paragraph{Impact of Categorization}  
Table~\ref{table:results} indicates that our category-aware model  outperforms the base model on both \aqua\ and MathQA datasets. The gain is relatively small because the current model only uses  categorization decisions as hard constraints at decoding time. Moreover, the problem categorization might be noisy due to   our use of only one mathematical interpretation for each domain-specific n-gram. For example, the presence of the words ``square" or ``cube" in the text of a math word problem indicate that the word problem is related to the geometry domain, but these unigrams can also refer to an exponential operation ($n^2$ or $n^3$).  

To measure the effectiveness of our categorization strategy, we used human annotation over 100 problems. The agreement between human annotators is $84\%$ and their agreement with our model is $74.5\%$. As a future extension of this work, we would like to also consider the context in which domain-specific n-grams appear.



\paragraph{Discussions}
As we mentioned in section \ref{sec:lang}, the continuous nature of our formalism allows us to solve problems requiring systems of equations. However, there are other types of word problems that are currently unsolvable or have multiple interpretations leading to multiple correct solutions. While problems that can only be solved by brute-force instead of logical reasoning and non-narrative problems that do not fit the definition of a math word problem (in Table \ref{table:solve-table} these appear as ``no word") are removed from consideration, there are other problems that are beyond the scope of current models but could pose an interesting challenge for future work. One example is the domain of sequence problems. Unlike past word problem-solving models, our models incorporate domain-specific math knowledge, which is potentially extensible to common sequence and series formulas.

%% file: conclusion.tex
\section{Conclusion}

In this work, we introduced a representation language and annotation system for large-scale math word problem-solving datasets that addresses unwanted noise in these datasets and lack of formal operation-based representations. We demonstrated the effectiveness of our representation language by transforming solvable \aqua\ word problems into operation formalisms. Experimental results show that both our base and category-aware sequence-to-program models outperform baselines and previous results on the \aqua\ dataset when trained on data aligned with our representation language. Our representation language provides an extra layer of supervision that can be used to reduce the influence of statistical bias in datasets like \aqua. Additionally, generated operation programs like the examples in figure \ref{fig:gen} demonstrate the effectiveness of these operation formalisms for representing math word problems in a human interpretable form. 

The gap between the performance of our models and human performance indicates that our \data\  still maintains the challenging nature of \aqua\ problems. In future work,  we plan to extend our representation language and models to cover currently unsolvable problems, including sequence and high-order polynomial problems. 

\paragraph{Acknowledgements} This research was supported by ONR (N00014-18-1-2826), NSF (IIS 1616112), Allen Distinguished Investigator Award, and gifts from Google, Allen Institute for AI, Amazon, and Bloomberg. We thank Marti A. Hearst, Katie Stasaski, and the anonymous reviewers for their helpful comments. 